\pdfoutput=1

\documentclass[11pt]{article}

\usepackage[final]{acl}

\usepackage{float}
\usepackage{times}
\usepackage{latexsym}
\usepackage[T1]{fontenc}
\usepackage[utf8]{inputenc}

\usepackage{microtype}

\usepackage{inconsolata}

\usepackage{graphicx}
\usepackage{amsmath}
\usepackage{multirow}
\usepackage{booktabs}
\usepackage{color}
\usepackage{hhline}
\usepackage[capitalize]{cleveref}
\usepackage{tcolorbox}
\usepackage{pifont}
\usepackage{subcaption}

\newcommand{\method}[0]{\textsc{AdaCAD}}

\title{\method{}: Adaptively Decoding to Balance Conflicts between \\ Contextual and Parametric Knowledge}
\author{Han Wang \;\;\;\;\; Archiki Prasad \;\;\;\;\; Elias Stengel-Eskin \;\;\;\;\;   Mohit Bansal \\
\textnormal{UNC Chapel Hill} \\ \texttt{\{hwang, archiki, esteng, mbansal\}@cs.unc.edu} \\ }

\begin{document}
\maketitle

\begin{abstract}
Knowledge conflict arises from discrepancies between information in the context of a large language model (LLM) and the knowledge stored in its parameters. 
This can hurt performance when using standard decoding techniques,
which tend to ignore the context.
Existing test-time contrastive methods seek to address this by comparing the LLM's output distribution with and without the context and adjust the model according to the contrast between them.
However, we find that these methods frequently misjudge the degree of conflict and struggle to handle instances that vary in their amount of conflict, with static methods over-adjusting when conflict is absent.  
We propose a fine-grained, instance-level approach called \method{}, which dynamically infers the weight of adjustment based on the degree of conflict, as measured by the Jensen-Shannon divergence between distributions representing contextual and parametric knowledge.
Across four LLMs, six question-answering (QA) and three summarization datasets, we demonstrate that \method{} consistently outperforms other decoding baselines with average QA accuracy gains of $14.21\%$ (absolute) over a static contrastive baseline, and improves the factuality of summaries by $6.19$ (AlignScore).
Lastly, we show that while contrastive baselines hurt performance when conflict is absent, \method{} mitigates these losses, making it more applicable to real-world datasets in which some examples have conflict and others do not.\footnote{Our code is publicly available at: \url{https://github.com/HanNight/AdaCAD}}  

\end{abstract}

\begin{figure}
    \centering
    \includegraphics[width=0.47\textwidth]{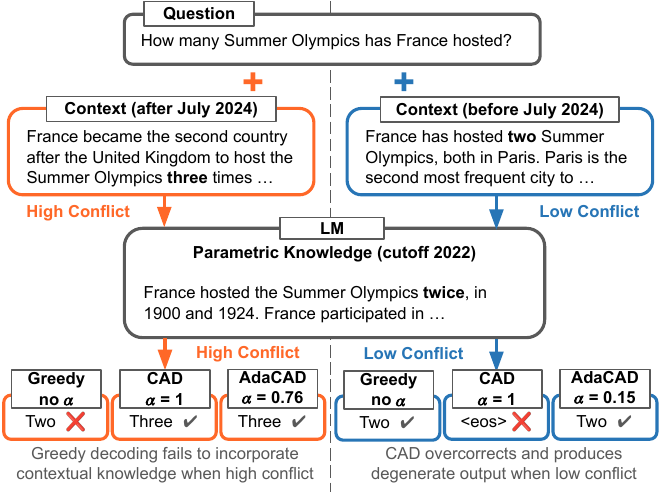}
    \caption{
    In cases of high knowledge conflict, greedy decoding fails to attend to the context, resulting in incorrect answers. Contrastive decoding allows the context to be incorporated, but must be done with care: in low-conflict cases, excessive contrast can over-correct (e.g., by CAD with $\alpha=1$), resulting in incorrect outputs. \method{} dynamically adjusts the degree of contrast, allowing it to handle both high and low-conflict cases.}
    \label{fig:fig_1}
\end{figure}

\section{Introduction}
Large language models (LLMs) encode vast amounts of information from pretraining in their parameters~\citep{petroni-etal-2019-language, roberts-etal-2020-much}, giving them remarkable capabilities in knowledge-intensive NLP tasks.
However, LLMs also hallucinate plausible but factually incorrect responses due to outdated knowledge \citep{lazaridou2021mind, dhingra-etal-2022-time, kasai2023realtime}, lesser-known facts~\citep{mallen-etal-2023-trust}, and even misinformation in the pre-training corpus. 
A popular line of prior work aims to improve answers and reduce hallucination by augmenting LLMs' context with external knowledge, including knowledge from retrieved documents \citep{pmlr-v119-guu20a, NEURIPS2020_6b493230}, web search results \citep{nakano2022webgpt}, and the outputs of tools \citep{schick2023toolformer}.
However, discrepancies between the added contextual knowledge and the model's pretrained parametric knowledge can cause \emph{knowledge conflict}.
In these cases, models often overlook the provided context and rely overly on the parametric knowledge~\cite{longpre-etal-2021-entity, chen-etal-2022-rich, zhou-etal-2023-context, wan-etal-2023-histalign}.
For example, in \cref{fig:fig_1}, the LLM's pretraining data (and thus its parametric knowledge) has a cutoff of September 2022, at which point France had hosted the Summer Olympics twice. 
This conflicts with the latest contextual knowledge (from July 2024) when France had hosted three times, and leads the model to answer incorrectly when using greedy decoding. 

One promising direction for handling knowledge conflict uses inference-time decoding strategies that adjust the model's output probability distribution without the need for additional training.
\citet{shi-etal-2024-trusting} propose context-aware decoding (CAD) which seeks to correct the model's output based on the difference between output probability distributions with and without the context. 
However, in practice, we find that while CAD works well when there is a uniformly high degree of conflict between the parametric knowledge and external context, it struggles with scenarios in which different examples have \emph{varying degrees of knowledge conflict}.  
Empirically, we observe that CAD can in fact degrade performance on low-conflict examples by \emph{overcorrecting} the output distribution. 
For example, in \cref{fig:fig_1}, when the context is sourced from a document before July 2024, there is no conflict between the parametric knowledge and the contextual knowledge; both state that France has hosted the Olympics twice. 
Here, CAD overcorrects the distribution, leading to an invalid answer.  

In this work, we present a simple and effective dynamic decoding method, \textbf{Ada}ptive \textbf{C}ontext \textbf{A}ware \textbf{D}ecoding (\method{})
, aimed at automatically modeling the degree of conflict between the context and parametric knowledge and dynamically inferring the degree of adjustment needed for every token. 
We use the Jensen-Shannon divergence (JSD) between output distributions with and without the context to measure the degree of knowledge conflict, using the resulting value to reweight the combination of distributions. 
A higher JSD indicates a greater degree of conflict and signals the need for higher adjustment (more weight on the contextual knowledge) while a lower JSD reflects a smaller degree of conflict requiring a smaller adjustment (more weight on the parametric knowledge). 
As illustrated in \cref{fig:fig_1}, this leads to correct answers \textit{both for high and low-conflict examples} by helping the model adaptively decide how to weigh contextual vs. parametric knowledge. 

We demonstrate \method{}'s effectiveness on a diverse range of tasks, covering question-answering (QA) and summarization, with six QA datasets (Natural Question (NQ; \citealp{kwiatkowski-etal-2019-natural}), \textsc{NQ-Swap} \citep{longpre-etal-2021-entity}, TriviaQA \citep{joshi-etal-2017-triviaqa}, PopQA \citep{mallen-etal-2023-trust}, HotpotQA \citep{yang-etal-2018-hotpotqa}, TabMWP \citep{lu2023dynamic}) and three summarization datasets (CNN-DM \citep{see-etal-2017-get}, XSum \citep{narayan-etal-2018-dont}, and TofuEval \citep{tang-etal-2024-tofueval}).
We test a range of base LLMs, examining Llama-2 \citep{touvron2023llama}, Llama-3 \citep{llama3modelcard}, and Mistral \citep{jiang2023mistral}. 
We consider different sizes of these models and also test both the base and instruction-tuned variants. 
Our results and analyses show that decoding with a uniform level of contrast benefits high-conflict scenarios but generally hurts performance,  while the adaptive contrast of \method{} results in improvements across the board.
Overall, \method{} generally achieves superior performance compared to the baselines, 
with an absolute gain of $14.21\%$ over CAD (a static baseline), $4.82\%$ over COIECD \citep[][a baseline that classifies instances as conflicting or not]{yuan-etal-2024-discerning}, $5.86\%$ over ConfCD \citep[][a method that makes dynamic token-level adjustments based on LLM confidence]{zhao-etal-2024-enhancing}, and $2.41\%$ over greedy decoding when averaged across models and QA datasets. 
On summarization, \method{} improves summary quality and factuality, with an average AlignScore \citep{zha-etal-2023-alignscore} gain of 4.16 over greedy decoding, 2.19 over CAD, 10.44 over COIECD, and 7.96 over ConfCD.

Furthermore, in our analyses, we explore \emph{why} \method{} improves over the baselines.
We first validate the hypothesis that \method{} is able to balance contextual and parametric knowledge by assigning lower weights to lower-conflict instances, testing each method on datasets designed to have high and low conflict, finding that \method{}'s inferred weight is much lower when there is no conflict. 
We also compare the amount by which CAD and \method{} adjust the base model's distribution on examples with and without conflict, finding that while \method{} changes the distribution less when there is no conflict (i.e., when the base model's distribution is already sufficient), CAD adjusts by roughly the same amount whether there is conflict or not, explaining its lower QA performance. 
Additionally, for summarization tasks, \method{} generates more faithful summaries whereas other methods tend to hallucinate details.

\section{Related Work}

\paragraph{Knowledge Conflict}
Integrating external knowledge as context into LLMs enables them to keep abreast of current world knowledge \citep{kasai2023realtime}, reduce hallucination \citep{shuster-etal-2021-retrieval-augmentation}, and improve factuality.
However, a recent line of work focuses on discrepancies between external contextual knowledge and the model's parametric knowledge, such as LLMs' over-reliance on their parametric knowledge on entity-based QA tasks~\cite{longpre-etal-2021-entity}, ignoring retrieved contexts~\cite{tan2024blinded}, and exhibiting confirmation bias~\cite{xie2024adaptive}, etc.
\citet{zhou-etal-2023-context} demonstrate that LLMs' faithfulness to the context can be significantly improved using carefully designed prompting strategies -- this is orthogonal to our work, which is compatible with different prompts. 
\citet{zhang-etal-2023-merging} address how to combine retrieved and parametric knowledge to improve open-domain QA, but require further training discriminators with silver labels, whereas our method is training-free. 

\paragraph{Contrast in Text Generation}
Contrastive approaches for text generation have been widely studied and used to enhance response diversity in conversations~\cite{li-etal-2016-diversity}, steering model generations towards desired attributes while maintaining fluency and diversity~\cite{liu-etal-2021-dexperts}, and contrasting between larger and relatively smaller language models to generate high-quality text~\cite{li-etal-2023-contrastive}, and improve visually-grounded generation tasks \citep{wan2024contrastive}. 
Context-aware decoding \citep[CAD;][]{shi-etal-2024-trusting} leverages a contrastive output distribution that amplifies the differences between the output probabilities predicted by a model with and without the context, promoting greater attention to the input context for more faithful and reliable text generation. 
Unlike \method{}, these past contrastive approaches do not adapt the weight on distributions to varying degrees of knowledge conflict. 
To address this, \citet{yuan-etal-2024-discerning} introduce COIECD, a decoding-time method that categorizes instances into two discrete bins -- high and low conflict -- based on a complex information-entropy constraint governed by tuned hyperparameters, and employs different decoding strategies (by altering CAD) for each. 
\citet{zhao-etal-2024-enhancing} uses LLM confidence to adjust the output probabilities dynamically (denoted as ConfCD) as well as relies on additional noisy and irrelevant contexts.
In contrast, \method{} employs a single dynamic instance-level strategy that automatically models (based on Jensen-Shannon divergence) a continuous degree of conflict without imposing rigid categories or requiring additional noisy and irrelevant contexts, accommodating more general knowledge conflict settings.
In addition to these conceptual differences, in \cref{sec:results}, we show that \method{} outperforms CAD, COIECD, and ConfCD on QA and summarization. 

\begin{figure*}[t]
    \centering
    \includegraphics[width=1\textwidth]{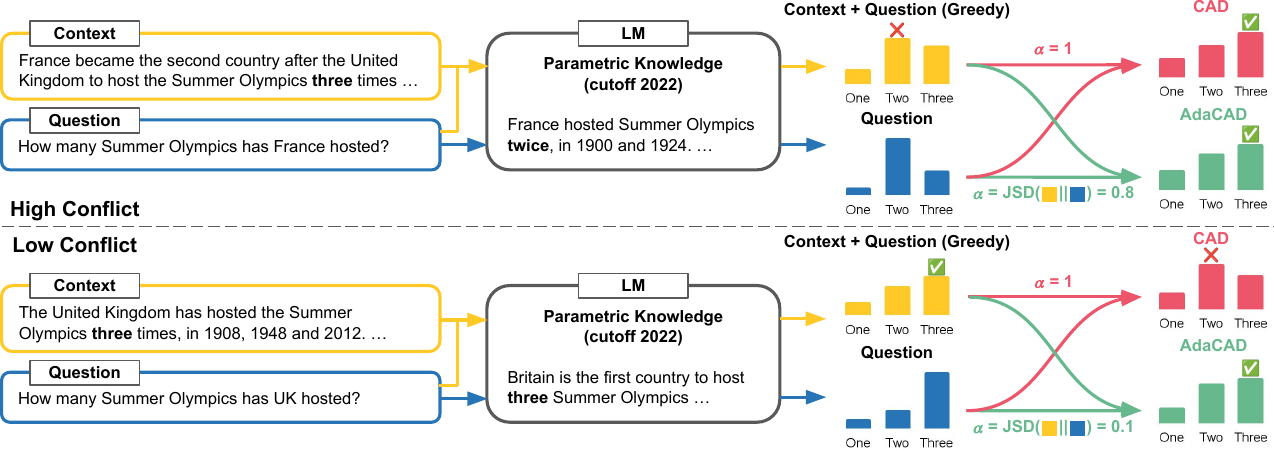}
    \caption{Comparison of greedy decoding (Context+Question), CAD, and \method{} on high-conflict and low-conflict examples. Greedy decoding struggles to incorporate contextual knowledge in high-conflict examples. 
    CAD tends to overemphasize irrelevant tokens in the vocabulary, leading to incorrect answers in low-conflict examples. 
    \method{} uses dynamic adaptation to effectively balance between context and parametric knowledge.
    }
    \label{fig:main_fig}
\end{figure*}

\section{Methodology}
\paragraph{Task and Notation}
Given an input query $\boldsymbol{x}$ with a relevant context $\boldsymbol{c}$, a language model parameterized by $\theta$ is tasked with generating a correct response $\boldsymbol{y} = y_1, \ldots, y_n$ of length $n$ that respects the context. At each decoding step $t$, a token $y_t$ can be sampled autoregressively from a probability distribution conditioned on query $\boldsymbol{x}$ and context $\boldsymbol{c}$ as $y \sim p_\theta(y \mid \boldsymbol{c}, \boldsymbol{x}, \boldsymbol{y}_{<t})$. 
However, when there is conflict between knowledge in the context $\boldsymbol{c}$ and parametric knowledge encoded in LLM, the model can struggle to pay enough attention to $\boldsymbol{c}$ and overly rely on the parametric knowledge \citep{longpre-etal-2021-entity, chen-etal-2022-rich}, i.e., sample from a distribution more akin to $p_\theta(y | \boldsymbol{x}, \boldsymbol{y}_{<t})$.

\paragraph{Background: Context-aware Decoding}
\label{ssec:cad}
To mitigate knowledge conflicts, \citet{shi-etal-2024-trusting} introduce Context-aware Decoding (CAD), which samples from a contrastive output distribution that amplifies the difference between output probabilities with and without context. 
CAD measures the parametric knowledge via $p_{\theta}(y | \boldsymbol{x}, \boldsymbol{y}_t)$ and prioritizes relevant contextual knowledge over the model’s parametric knowledge by using the pointwise mutual information (PMI) between the context $\boldsymbol{c}$ and the generation $y$, conditioned on $\boldsymbol{x}, \boldsymbol{y}_{<t}$ to modify the model's original output distribution. 
\begin{align}
    y_t &\sim \Tilde{p}_{\theta}\left(y \mid \boldsymbol{c}, \boldsymbol{x}, \boldsymbol{y}_{<t}\right) \nonumber \\
    & \propto p_{\theta}(y \mid \boldsymbol{c}, \boldsymbol{x}, \boldsymbol{y}_{<t})\!\left[\frac{p_{\theta}(y \mid \boldsymbol{c}, \boldsymbol{x}, \boldsymbol{y}_{<t})}{p_{\theta}(y \mid \boldsymbol{x}, \boldsymbol{y}_{<t})}\right]^{\alpha} \label{eqn:main}
\end{align}
where the PMI term $\frac{p_{\theta}\left(y \mid \boldsymbol{c}, \boldsymbol{x}, \boldsymbol{y}_{<t}\right)}{p_{\theta}\left(y \mid \boldsymbol{x}, \boldsymbol{y}_{<t}\right)}$ is a scaling factor used to adjust the parametric knowledge, and $\alpha$ governs the weight or degree of adjustment.
A larger $\alpha$ means a greater adjustment and $\alpha = 0$ reduces to no adjustment, i.e., greedy decoding.\footnote{While PMI also measures the amount of conflict between the distributions with and without context, empirically, we find that it still results in a high degree of perturbation to the output distribution in cases of low conflict (c.f. \cref{ssec:pmi}).}

\paragraph{\method{}: Handling Variable Conflict}
\label{ssec:adacod}
In test-time contrastive methods -- such as those presented by \citet{li-etal-2023-contrastive} and \citet{shi-etal-2024-trusting} --  $\alpha$ is a fixed hyperparameter set for an entire dataset, requiring tuning on a validation set.
However, every instance in the dataset may need a different weight for adjustment; furthermore, in longer-form generation, individual timesteps may require different weights, making a single $\alpha$ value suboptimal. 
For instance, in the presence of a high degree of conflict, e.g., \cref{fig:main_fig}~(top), a larger $\alpha$ can perturb the LLM's output distribution to mitigate over-reliance on parametric knowledge, whereas in cases with low or no conflict (as in \cref{fig:main_fig}~(bottom)), the adjustment to the LLM's output distribution is minimal.
Therefore, a fixed $\alpha$ may fail on scenarios where there are heterogeneous examples with and without conflict, i.e., on realistic datasets. 

To address variable conflict, we introduce a different $\alpha_t$ for each timestep and each instance.
Specifically, we automatically infer $\alpha_t$ dynamically based on the degree of knowledge conflict for each instance (and decoding step) without supervision, enabling automatic adaptation.
To accomplish this, we use Jensen-Shannon divergence \citep[JSD;][]{lin1991divergence} to model the degree of conflict between the context and parametric knowledge. 
While similar to Kullback-Leibler divergence, JSD is symmetric and bounded within the range $[0, 1]$, making it more suitable for modeling conflicts, as it provides a more interpretable and normalized measure of divergence (details in \cref{app:jsd}).
A larger JSD between $p_\theta(y \mid \boldsymbol{x}, \boldsymbol{y}_t)$ and $p_\theta(y \mid \boldsymbol{c}, \boldsymbol{x}, \boldsymbol{y}_t)$ reflects a greater conflict between context and parameter knowledge, suggesting that we need a larger $\alpha$ to encourage the LM to rely more on the context, while a smaller JSD reflects a smaller conflict, suggesting that a smaller $\alpha$ is required to maintain the LM's adherence to its parametric knowledge.
Therefore, we set $\alpha^{\textsc{jsd}}_t$ at each decoding step $t$ to:
\begin{equation*}
    \alpha^{\textsc{jsd}}_t = \mathrm{JSD}\left(p_{\theta}\left(y_t \mid \boldsymbol{x}, \boldsymbol{y}_{<t}\right)\parallel p_{\theta}\left(y_t \mid \boldsymbol{c}, \boldsymbol{x}, \boldsymbol{y}_{<t}\right)\right)
\end{equation*}
This enables both coarse-grained instance-level and fine-grained token-level adjustments.
Finally, we sample outputs from the probability distribution:
\begin{equation*}
    y_t \sim p_{\theta}(y\!\mid\!\boldsymbol{c}, \boldsymbol{x}, \boldsymbol{y}_{<t})\!\left[\frac{p_{\theta}(y \!\mid\! \boldsymbol{c}, \boldsymbol{x}, \boldsymbol{y}_{<t})}{p_{\theta}(y\!\mid\!\boldsymbol{x}, \boldsymbol{y}_{<t})}\right]^{\!\alpha^{\textsc{jsd}}_t}
\end{equation*}

This dynamic adaptation allows our approach to effectively balance between context and parametric knowledge, ensuring robust performance across varying degrees of conflict without the need for extensive manual tuning, thereby enhancing both flexibility and accuracy in diverse scenarios.

\paragraph{\method{} for Long-form Generation}
In long-form generation tasks, we find that initially, the JSD values tend to be low (cf. \cref{fig:jsd_trend} in \cref{app:jsd_trend}). 
This may be due to the model's tendency to produce generic, low-information outputs at the start of each sequence. 
Therefore, the divergence between $p_\theta(y_t | \boldsymbol{x}, \boldsymbol{y}_{<t})$ and $p_\theta(y_t | \boldsymbol{c}, \boldsymbol{x}, \boldsymbol{y}_{<t})$ is minimal. 
To mitigate this issue and ensure more consistent performance throughout the generation process, we introduce a warmup operation: 
$\alpha^{\textsc{jsd}}_t = \max\left(\alpha^{\textsc{jsd}}_t, \lambda\right)$,
where $\lambda$ is a lower bound to adjust for the initially low JSD values, ensuring a more robust and stable starting point. We set $\lambda = 0.3$ for long-form generation tasks.\footnote{We set $\lambda = 0.3$ to match the maximum JSD values for non-conflicting data from QA (cf. \cref{ssec:analysis}).} 

\section{Experiments and Results}
\subsection{Experimental Setup} 
\label{sec:setup}

\paragraph{Datasets and Metrics}
We evaluate on several QA datasets: Natural Questions (NQ; \citealp{kwiatkowski-etal-2019-natural}), TriviaQA \citep{joshi-etal-2017-triviaqa}, PopQA \citep{mallen-etal-2023-trust}, and HotpotQA \citep{yang-etal-2018-hotpotqa}. 
We use these datasets to simulate real scenarios with varying degrees of conflict for each instance.
Additionally, we evaluate on an existing knowledge conflict dataset, \textsc{NQ-Swap} \citep{longpre-etal-2021-entity}, which is based on the NQ dataset and consists of synthetic conflicting data. 
Lastly, we also test on a popular tabular question-answering dataset, TabMWP \citep{lu2023dynamic}, that requires LLMs to use reasoning skills over tabular contexts.
We report exact match accuracy on all QA datasets. 

\begin{table*}[t]
\small
\centering
\begin{tabular}{llccccccc}
\toprule
\multicolumn{1}{l}{\textbf{Model}} & \textbf{Decoding} & \textbf{NQ }& \textbf{\textsc{NQ-Swap}} & \textbf{TriviaQA} & \textbf{PopQA} &\textbf{ HotpotQA }& \textbf{TabMWP} & \textbf{Avg} \\
\midrule
\multirow{5}{*}{Llama2-13B} & Greedy & 44.26 & 54.89 & 85.50 & 76.65 & 38.27 & 38.30 & 56.31 \\
 & CAD & 37.91 & \textbf{80.35} & 71.40 & 76.83 & 31.92 & 19.30 & 52.95 \\
 & COIECD & 44.60 & 59.84 & \textbf{87.00} & \textbf{81.05} & \textbf{42.81} & \textbf{38.80} & \textbf{59.02}  \\
 & ConfCD & 45.81 & 76.89 & 81.70 & 79.08 & 35.11 & 29.10 & 57.95 \\
 & \method{} & \textbf{46.73} & 67.84 & 85.40 & 78.79 & 37.83 & 37.50 & \textbf{59.02} \\
\midrule
\multirow{5}{*}{Llama3-8B} & Greedy & 44.63 & 47.81 & \textbf{85.70} & 80.51 & \textbf{51.42} & 52.20 & 60.38 \\
 & CAD & 35.96 & \textbf{77.94} & 40.20 & 74.27 & 39.53 & 26.60 & 49.08 \\
 & COIECD & 43.36 & 51.16 & 83.10 & 78.49 & 45.63 & 49.70 & 58.57 \\
 & ConfCD & 42.90 & 72.44 & 71.20 & 79.80 & 47.13 & 46.20 & 59.95 \\
 & \method{} & \textbf{45.47} & 62.34 & 82.50 & \textbf{81.34} & 50.53 & \textbf{53.00} & \textbf{62.53} \\
\midrule
\multirow{5}{*}{Llama3-70B} & Greedy & 44.13 & 55.74 & \textbf{90.20} & \textbf{86.10} & \textbf{56.11} & 66.70 & 66.50 \\
 & CAD & 34.05 & \textbf{81.32} & 54.60 & 75.16 & 40.86 & 48.60 & 55.77 \\
 & COIECD & 45.09 & 57.26 & 88.60 & 83.60 & 52.03 & 64.40 & 65.16 \\
 & ConfCD & 41.44 & 79.34 & 81.00 & 82.00 & 50.14 & 63.50 & 66.24 \\
 & \method{} & \textbf{45.43} & 70.07 & 88.80 & 85.68 & 55.00 & \textbf{67.10} & \textbf{68.68} \\
\midrule
\multirow{5}{*}{Mistral-7B} & Greedy & 42.56 & 56.86 & \textbf{80.40} & \textbf{67.56} & 40.89 & 38.90 & 57.65 \\
 & CAD & 20.98 & 66.89 & 24.20 & 48.54 & 18.49 & 20.10 & 35.82 \\
 & COIECD & 29.00 & 58.09 & 71.60 & 64.59 & 35.83 & 31.60 & 48.45 \\
 & ConfCD & 23.99 & 59.29 & 58.70 & 54.19 & 29.83 & 31.30 & 42.88 \\
 & \method{} & \textbf{45.09} & \textbf{67.27} & 80.20 & 67.26 & \textbf{41.35} & \textbf{39.70} & \textbf{60.23} \\
 \bottomrule
\end{tabular}
\caption{Under a zero-shot setting, we show that on average (across tasks and models) \method{} improves accuracy by $14.21\%$ over CAD, $4.82\%$ over COIECD, and $5.86\%$ over ConfCD (results with instruction-tuned models in \cref{app:post-train}).}
\label{tab:base_qa_0shot_res}
\end{table*}

To test \method{} on longer-form generation tasks, we evaluate on three standard summarization tasks: CNN-DM \citep{see-etal-2017-get}, XSum \citep{narayan-etal-2018-dont}, and TofuEval \citep{tang-etal-2024-tofueval}.
While many documents from older datasets (such as CNN-DM and XSum) are present in LLM's pretraining data,\footnote{
Using \url{pile.dataportraits.org}, we find several documents from CNN-DM appear in the Pile \citep{gao2020pile}, commonly used to pretrain LLMs.
} TofuEval is a recent, more challenging benchmark on topic-focused dialogue summarization (especially for marginal or secondary topics in the document).
We use two reference-based metrics, ROUGE-L \citep{lin-2004-rouge} and BERT-P \citep{Zhang2020BERTScore}, to evaluate summarization quality. 
As TofuEval does not support reference-based evaluation~\citet{tang-etal-2024-tofueval}, we use recommended
AlignScore \citep{zha-etal-2023-alignscore} to measure the factual consistency of summaries on both \emph{main} (central to the document) and \emph{marginal} (lesser explored) topics.
For additional details and examples of all datasets, refer to \cref{app:data}. 

\paragraph{Source of Context}
We use the gold context provided by NQ, \textsc{NQ-Swap}, TriviaQA, and HotpotQA as the relevant contexts.
Since PopQA does not provide gold contexts, we employ BM25 \citep{Robertson2009ThePR}, to retrieve relevant contexts from Wikipedia.
For TabMWP, we take the semi-structured table as the relevant context.
In summarization tasks, the source document serves as the relevant context, while the instruction is used as the input query.
A summary of input query $\boldsymbol{x}$ and context $\boldsymbol{c}$ for all datasets is shown in \cref{tab:input_example}  with corresponding prompts in \cref{app:prompts}.

\paragraph{Models}
We test \method{} on different pre-trained base language models, including Llama2 (13B) \citep{touvron2023llama}, Llama3 (8B, 70B) \citep{llama3modelcard}, and Mistral (7B) \citep{jiang2023mistral};
we measure \method{}'s effectiveness both on the base and instruction-tuned model variants.

\begin{table*}[t]
\small
\centering
\hspace*{-0.35cm}
\begin{tabular}{lccc|ccc|cc}
\toprule
 & \multicolumn{3}{c}{\textbf{CNN-DM}} & \multicolumn{3}{c}{\textbf{XSum}} & \multicolumn{2}{c}{\textbf{TofuEval (AlignScore)}} \\ \cmidrule{2-9} 
\textbf{Decoding} & \textbf{ROUGE-L} & \textbf{BERT-P} & \textbf{AlignScore} & \textbf{ROUGE-L} & \textbf{BERT-P} & \textbf{AlignScore} & \textbf{Overall} & \textbf{Main / Marginal} \\ \midrule
Greedy & 24.93 & \textbf{95.41} & 91.44 & 14.36 & 94.05 & 85.28 & 76.66 & 81.64 / 61.19 \\
CAD & 24.76 & 94.45 & 91.01 & 14.59 & 93.65 & 84.34 & 83.93 & 87.26 / 73.58 \\
COIECD & 23.47 & 92.06 & 85.49 & 14.51 & 91.04 & 73.81 & 75.24 & 80.68 / 58.31 \\
ConfCD & 23.94 & 93.37 & 87.03 & 14.78 & 92.71 & 77.98 & 76.97 & 78.17 / 73.23 \\
\method{} & \textbf{25.42} & 94.91 & \textbf{94.97} & \textbf{14.91} & \textbf{94.29} & \textbf{85.81} & \textbf{85.07} & \textbf{88.06} / \textbf{75.79} \\ \bottomrule
\end{tabular}
\caption{Results on summarization datasets with Llama3-70B showing \method{} yields the best performance on factuality metrics (AlignScore) and overall summarization quality (ROUGE-L, and BERT-P). The full results with other language models are shown in \cref{tab:sum_res_full} of \cref{app:sum_res_full}.}
\label{tab:sum_res}
\end{table*}

\paragraph{Baselines}
We compare \method{} to standard decoding,  context-aware decoding \citep[CAD;][]{shi-etal-2024-trusting} -- which has a fixed $\alpha$, COIECD~\citep{yuan-etal-2024-discerning} -- which classifies whether there is knowledge conflict using a method controlled by tuned thresholds and then operates in two different decoding modes, each with the same fixed $\alpha$, and ConfCD~\citep{zhao-etal-2024-enhancing} -- which dynamically sets alpha based on LLM confidence.
Across all tasks and baselines, we use greedy decoding under a zero-shot setting.\footnote{We find  that greedy decoding outperforms top-$p$ sampling on CNN-DM, so we use greedy decoding across all methods for summarization tasks instead of top-$p$ sampling as in \citet{shi-etal-2024-trusting} (c.f. \cref{tab:diff_decode}, \cref{app:diff_decode}).}
For CAD, we set $\alpha = 1$ for the QA datasets and $\alpha = 0.5$ for the summarization datasets, following prior work \cite{shi-etal-2024-trusting}.
For COIECD, the values of $\lambda$ and $\alpha$ are set to 0.25 and 1 for QA datasets, and 0.25 and 0.5 for the summarization datasets, respectively, following \citet{yuan-etal-2024-discerning}.
For ConfCD, the $\alpha$ values are set to the maximum token probability with context ($C_R = \max_{y' \in V}p_\theta(y' | \boldsymbol{c}, \boldsymbol{x}, \boldsymbol{y}_{<t})$) if $C_R$
exceeds the maximum token probability without
context (i.e. $C_R > C = \max_{y' \in V}p_\theta(y' | \boldsymbol{x}, \boldsymbol{y}_{<t})$); otherwise, it is given by $1 - C$.
In \method{}, the $\alpha$ values are dynamically adjusted based on the degree of knowledge conflict for each instance. 

\subsection{Main Results}
\label{sec:results}

\paragraph{QA Tasks}
From \cref{tab:base_qa_0shot_res}, we observe that \method{} \emph{consistently} outperforms greedy decoding, CAD, COIECD, and ConfCD.
For instance, on Llama3-70B, \method{} achieves an average score improvement of $2.18\%$ (absolute) over greedy decoding, $12.91\%$ over CAD, $3.52\%$ over COIECD, and $2.44\%$ over ConfCD.
Note that while CAD performs quite well on \textsc{NQ-Swap} (containing \emph{only} high-conflict examples), it often degrades performance (relative to greedy decoding) on other QA datasets, resulting in an $18.58\%$ accuracy drop on average across all models and tasks; in contrast, \method{} performs well across datasets, whether they have conflict or not.
Furthermore, \method{} consistently outperforms COIECD across various QA datasets, highlighting the strength of our continuous JSD-based approach over COIECD's binary classification approach that splits instances into ones with conflict or without. 
For instance, \method{} outperforms COIECD by a large average margin of $10.29\%$ on NQ-SWAP across all models.
Additionally, on more complex datasets like TabMWP with newer LLMs, \method{} also shows superior performance against all baselines, e.g., achieving average improvements of $6.30\%$ with Llama3-70B and $9.23\%$ with Mistral-7B.
These results indicate that \method{} is better able to combine the advantages of greedy decoding and CAD, performing well in scenarios without knowledge conflict (as greedy decoding does) as well as those with conflict (as CAD does).


\paragraph{Summarization Tasks}
In \cref{tab:sum_res}, we investigate how \method{} can improve performance on longer-form generation, showing results on three summarization tasks, CNN-DM, Xsum, and TofuEval. 
For TofuEval, \method{} demonstrates substantial improvements, particularly excelling in marginal topics (i.e., topics not central to the document) where it outperforms greedy decoding, CAD, COIECD, and ConfCD by 14.60, 2.21, 17.48, and 2.56 points in terms of AlignScore -- a measure of faithfulness -- respectively.
This highlights \method{}'s ability to handle diverse topics and maintain factual consistency, especially when prompted to focus on a marginal topic; qualitatively, we see in \cref{fig:summ_exp} that these improvements are driven by less hallucination on the part of \method{}.

On CNN-DM, \method{} achieves the highest ROUGE-L score of 25.42, surpassing greedy decoding, CAD, COIECD, ConfCD by 0.49, 0.66, 1.95, and 1.48, respectively.
In terms of factual consistency, \method{} also leads with an AlignScore of 94.97.
On XSum, \method{} also outperforms all baselines across all metrics. 
For instance, \method{} achieves an average improvement of 1.43, and 5.46 points in BERT-P, and AlignScore, respectively. 
For BERT-P metric on CNN-DM, \method{} outperforms all contrastive decoding baselines and is slightly lower than Greedy decoding; as mentioned in \cref{sec:setup}, this may be a result of a lack of conflict in these datasets, which are at least partly included in large pretraining corpora. 
These improvements indicate that \method{}'s dynamic adjustment mechanism is effective in long-form generation, allowing it to balance context and parametric knowledge.

\section{Analysis}
\subsection{Performance comparison on instances with higher and lower degrees of conflict}
\label{ssec:analysis}

\paragraph{Setup}
In \cref{tab:base_qa_0shot_res}, we find that CAD underperforms \method{}, COIECD, as well as greedy decoding on most QA datasets, except \textsc{NQ-Swap}, wherein every instance by design has a high degree of conflict~\cite{longpre-etal-2021-entity}.
We hypothesize that on more realistic datasets, the trailing performance of CAD stems from its inability to account for instances with low or minimal conflict. To test this hypothesis, we evaluate all methods on examples designed to have \emph{minimal} conflict, i.e., where the model's internal representation aligns well with the context. 
Specifically, we generate a dataset of synthetic non-conflicting data called \textsc{NQ-Synth}: we sample 500 questions from Natural Questions and then prompt the Llama-3-70B to generate the answer for each question.
We replace the gold answer entity in the context with the generated answer by regex, thus, making the context consistent with the LLM's internal knowledge. 
Finally, we evaluate Llama-3-70B on \textsc{NQ-Synth} and on \textsc{NQ-Swap}. 
See \cref{tab:nq_swap_example} in \cref{app:data} for examples of \textsc{NQ-Swap} and \textsc{NQ-Synth}.

\begin{table}[t]
   \small
    \centering
    \begin{tabular}{lccc}
    \toprule
    \textbf{Decoding} & \textbf{\textsc{NQ-Swap}} & \textbf{\textsc{NQ-Synth}} & \textbf{Overall} \\ 
    \midrule
    Greedy & 51.60 & 88.20 & 69.90 \\
    CAD & 79.60 & 64.00 & 71.80 \\
    COIECD & 50.80 & 83.60 & 67.20 \\
    \method{} & 62.80 & 86.40 & \textbf{74.60} \\
    \bottomrule
    \end{tabular}
    \caption{Accuracy on conflicting data (\textsc{NQ-Swap}) and non-conflicting data (\textsc{NQ-Synth}) with Llama3-70B.}
    \label{tab:synthetic}
\end{table}

\paragraph{Result: CAD hurts performance when conflict is low, while \method{} can handle both cases.}
Consistent with our hypothesis, in \cref{tab:synthetic}, we observe that in the absence of conflict (on \textsc{NQ-Synth}), CAD substantially degrades performance by $\approx\!24\%$ relative to greedy decoding, while \method{} maintains a comparable performance.
Although COIECD seeks to detect conflict and operates in two distinct decoding modes for high and low confllict, it also underperforms in non-conflict scenarios, falling $2.8\%$ behind \method{}. However, in cases of high conflict (\textsc{NQ-Swap}), where greedy decoding yields dramatically lower accuracy, ~\method{} improves over greedy decoding by $11.2\%$, while COIECD cannot handle high-conflict examples as well, lagging behind \method{} by $12\%$. To further investigate how \method{} balances instances with lower and higher degrees of conflict, we compute $\alpha^{\textsc{jsd}}_{\mathrm{max}}$, which is the maximum $\alpha^{\textsc{jsd}}_t$ value across tokens, for both datasets. Indeed, we find that $\alpha^{\textsc{jsd}}_{\mathrm{max}}$ adapts to the amount of conflict, with an average value of $0.45$ on \textsc{NQ-Swap} with a higher level of conflict, and substantially lower value ($\alpha^{\textsc{jsd}}_{\mathrm{max}}\!=\!0.28$) on \textsc{NQ-Synth} which does not contain any conflict by design.

\subsection{PMI does not adequately address conflict}
\label{ssec:pmi}
As described in \cref{ssec:cad,ssec:adacod}, both CAD and \method{} compute the PMI between the LLM's output distributions with and without external context $\boldsymbol{c}$. However, CAD relies solely on the PMI term to balance the level of conflict, whereas in \method{}, we compute $(\text{PMI})^{\alpha_t^{\textsc{jsd}}}$ where both PMI and $\alpha_t^{\textsc{jsd}}$ adapt with the degree of conflict. 
In cases of low conflict, the LLM's distributions should in principle be the same with and without context, rendering $\text{PMI}\!\approx\!1$, i.e., resorting to greedy decoding for any value of $\alpha$ (cf. \cref{eqn:main}). However, in practice, we find that, even with minimal conflict, the PMI term reranks the tokens in the head of the LLM's distribution, resulting in poor performance for CAD.

\begin{table}[t]
\small
\centering
\begin{tabular}{lccc}
\toprule
\textbf{Decoding} & \textbf{$\boldsymbol{\rho}$ (\textsc{NQ-Swap})} & \textbf{$\boldsymbol{\rho}$ (\textsc{NQ-Synth})} & $\lvert\Delta\boldsymbol{\rho}\rvert$ \\ \midrule
CAD & 0.56 & 0.57 & 0.01 \\
\method{} & 0.86 & 0.94 & 0.08 \\
\bottomrule
\end{tabular}
\caption{Spearman rank-order correlation coefficient between original and adjusted output distributions for CAD and \method{} on \textsc{NQ-Swap} and \textsc{NQ-Synth}. The difference  $\lvert\Delta\boldsymbol{\rho}\rvert$ measures the sensitivity of a decoding method to the degree of conflict (higher is better).}
\label{tab:rank_cor}
\end{table}
    
\paragraph{Setup}
To test how well the PMI term accounts for conflict, we measure the amount of reranking (among tokens) done by CAD and \method{} relative to the greedy distribution. 
We compute the Spearman rank-order correlation coefficient $\rho$ between the greedy distribution and output distribution from CAD and \method{} (with scaling factors PMI and (PMI)$^{\alpha_t^{\textsc{jsd}}}$ respectively). 
We restrict the measurement to the top-20 tokens (averaged across decoding steps) on \textsc{NQ-Swap} and \textsc{NQ-Synth}.\footnote{As the rank of low-probability tokens does not influence the generation, we focus on the top-20 tokens at each step.}
Intuitively, a method \textit{sensitive to the degree of conflict} should yield a lower rank correlation (more perturbation) when the amount of conflict is high (on \textsc{NQ-Swap}), and higher rank correlation (less perturbation) in cases of low conflict (on \textsc{NQ-Synth}). 
To this end, we compute the absolute difference or \textit{sensitivity}, $\lvert\Delta\boldsymbol{\rho}\rvert$ between the two $\boldsymbol{\rho}$ values of \textsc{NQ-Swap} and \textsc{NQ-Synth}.
A larger $\lvert\Delta\boldsymbol{\rho}\rvert$ indicates that the method is more effective at distinguishing between conflicting and non-conflicting data, i.e., more sensitive to the degree of conflict in instances.

\begin{table}[t]
\small
\centering
\begin{tabular}{lcc}
\toprule
\textbf{Datasets} & CAD (tuned $\alpha$) & \method{} \\ \midrule
\textbf{NQ} & 44.35 (0.25) & \textbf{45.47} \\
\textbf{TriviaQA} & 79.60 (0.25) & \textbf{82.50} \\
\textbf{PopQA} & 78.19 (0.25) & \textbf{81.34} \\
\textbf{HotpotQA} & 46.81 (0.50) & \textbf{50.53} \\
\textbf{TabMWP}	& 46.90 (0.50) & \textbf{53.00} \\ \midrule
\textbf{Average} & 59.17 & \textbf{62.57} \\
\bottomrule
\end{tabular}
\caption{Performance of CAD with tuned $\alpha$ and \method{} on QA datasets with Llama3-8B.}
\label{tab:tune_alpha}
\end{table}

\begin{figure*}[t]
    \centering
    \includegraphics[width=1\textwidth]{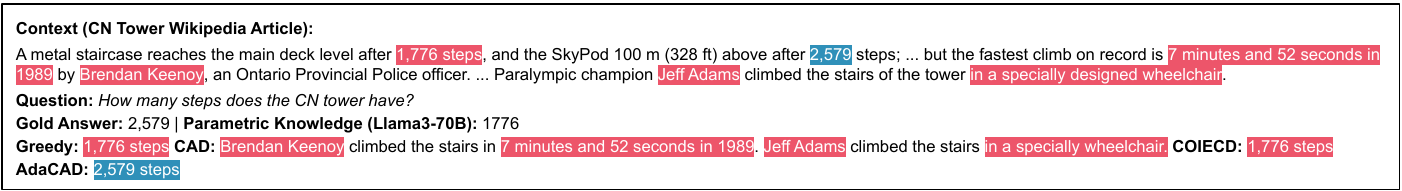}
    \caption{Qualitative example from NQ. Informative text is highlighted in blue, while text about unrelated facts and wrong answers is highlighted in red. \method{} produces the correct answer, while CAD generates unrelated outputs and COIECD fails to detect the conflict and generates the same incorrect answer as greedy decoding.}
    \label{tab:qual}
\end{figure*}

\begin{figure*}[t]
    \centering
    \includegraphics[width=1\textwidth]{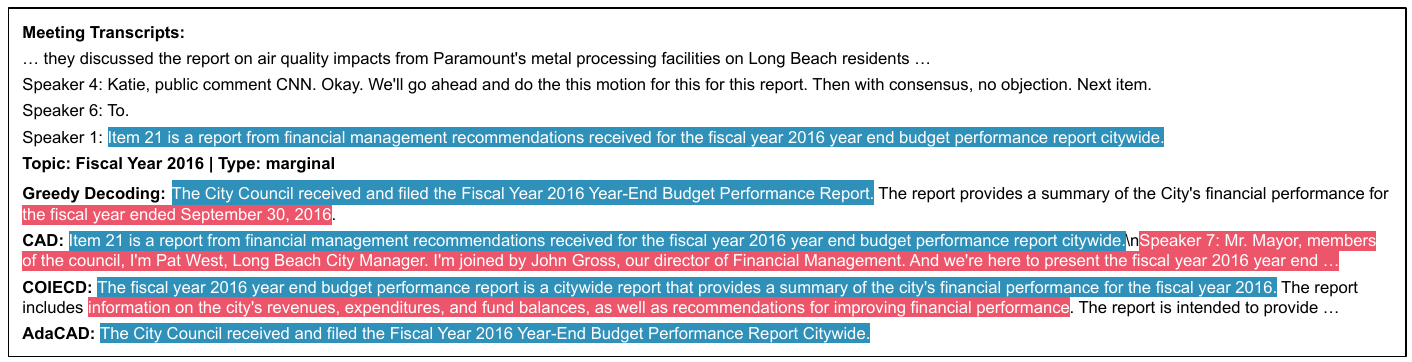}
    \caption{TofuEval: Text unsupported by the transcripts is highlighted in red, while consistent and relevant text is highlighted in blue. \method{} generates a faithful summary whereas other methods tend to hallucinate details.
    }
    \label{fig:summ_exp}
\end{figure*}

\paragraph{Result: PMI over-perturbs greedy distribution in low conflict setting; \method{} is adaptive.}
Results in \cref{tab:rank_cor} demonstrate that CAD, which only relies on the PMI term to offset conflicts, perturbs the greedy distribution to roughly the same extent ($\rho$) in the presence or absence of conflict, i.e., on \textsc{NQ-Swap} and \textsc{NQ-Synth}, respectively.  This minimal difference in $\lvert\Delta\boldsymbol{\rho}\rvert$ suggests that CAD is agnostic to the amount of conflict, leading to over-correction for non-conflicting examples. 
On the other hand, the correlation coefficient of \method{} is higher on \textsc{NQ-Synth} than on \textsc{NQ-Swap} (0.94 vs. 0.86), indicating more perturbation to the greedy distribution in the presence of conflict. 
Additionally, the sensitivity to conflict ($\lvert\Delta\boldsymbol{\rho}\rvert$) of \method{} is substantially larger ($8\times$) than that of CAD, highlighting \method{}'s superior ability to distinguish between conflicting and non-conflicting examples. 
Note that \method{} has a higher $\rho$ in both settings, indicating that overall, it perturbs the LLM's distribution to a lesser extent. 

\subsection{Tuning $\alpha$ of CAD for each dataset}

Since \method{} does not require validation data to tune the value of $\alpha$, we set CAD's $\alpha = 1$ (tuned on NQ-SWAP) for QA datasets following \citet{shi-etal-2024-trusting}, which may explain the strong performance of CAD on NQ-SWAP and low performance on other datasets.
To further underscore the advantages of \method{} over CAD, we compare \method{} (untuned) to a CAD baseline with a tuned $\alpha$ value. 
Specifically, we tune CAD's $\alpha$ using a validation set of 500 instances (randomly sampled from the train set) for each dataset.

\cref{tab:tune_alpha} shows that \method{} achieves an average improvement of $3.4\%$ (absolute) over CAD even when $\alpha$ is tuned.
We hypothesize that \method{}’s superior performance stems from varying the level of adjustment adaptively depending on the \textit{underlying instance}, whereas a tuned-$\alpha$ CAD still uses the same $\alpha$ uniformly for all instances and does not adjust according to varying degrees of conflict among instances. 
Moreover, while tuning CAD’s $\alpha$ for each dataset might improve performance in a controlled setting, such tuning does not scale well to real-world scenarios wherein models encounter a mix of user queries – some with high conflict and others with low or no conflict – and this categorization is not known a priori.

\subsection{Qualitative Examples}

\paragraph{QA Tasks}
We highlight the importance of adaptively capturing the degree of conflict between the context and parametric knowledge, we show a qualitative example from the NQ dataset in \cref{tab:qual}.
\method{} outperforms other methods by correctly generating the ``\textit{CN Tower's 2,579 steps}''.
We also observe that CAD tends to produce unrelated outputs due to over-correction, which over-amplifies the influence of irrelevant tokens within the vocabulary.
COIECD fails to detect the conflict and generates the same incorrect answer as greedy decoding.
We find that \method{} strikes the best balance between providing contrast in scenarios with high knowledge conflict while not suffering from over-correction on low-conflict instances. 

\paragraph{Summarization}
We also show a qualitative example from TofuEval in \cref{fig:summ_exp}. Given a meeting transcript centered on \emph{``report on air quality impacts from Paramount's metal processing facilities on Long Beach residents''}, we prompt the LLMs to generate a summary focused on the marginal topic \emph{``Fiscal Year 2016''}.
Baselines like greedy decoding, CAD, and COIECD tend to hallucinate details, such as fabricated financial data or names of individuals not mentioned in the transcript, which are highlighted in red.
In contrast, \method{} generates a more accurate and faithful summary without introducing unverified information.

\section{Discussion and Conclusion}
In naturalistic scenarios with mixed datasets containing examples with and without knowledge conflicts, existing decoding methods, including CAD, fail to adapt to changing amounts of conflict and in fact can lead to \emph{reduced} performance. 
Although larger and more performant models can store more information in their parametric knowledge -- thus leading to less and less conflict as models improve -- there will still always be gaps between the model and the actual state of the world (e.g., because of time cutoffs).
This means that models will encounter both low- and high-conflict scenarios, no matter their strength.  

To this end, we introduce \method{}, a simple yet effective dynamic decoding method that uses Jensen-Shannon divergence to dynamically model the degree of conflict for a given example (and timestep) and automatically balance the contrast between contextual and parametric knowledge. 
On diverse QA datasets, we show that \method{} combines the best of greedy decoding and context-aware decoding, improving performance. Additionally, experiments on summarization demonstrate that \method{} enhances both the quality and factuality of generated text, while other methods tend to hallucinate details.
Lastly, \method{} consistently outperforms COIECD, another hybrid decoding strategies that detects conflict.
Our analysis reveals that \method{} mitigates the overcorrection seen in CAD by dynamically adjusting the weight of contextual knowledge based on the degree of conflict.

\section*{Limitations}
Since our proposed method \method{} is based on CAD, it requires access to output logits from LLMs to calculate the difference between output probabilities with and without context. 
However, API-based LLMs like GPT-4 often do not provide output logits, making it challenging to directly apply logit-based methods like \method{} and CAD to fully black-box models.
Additionally, our experiments focus on English datasets and pre-trained models; as LLMs become available for other languages, future research will be needed to explore the interactions between language and knowledge conflict.
We do not foresee any particular risks associated with the application of our method.

\section*{Acknowledgements}
We would like to thank David Wan for feedback on our summarization experiments and the anonymous reviewers for their feedback.  
This work was supported by DARPA ECOLE Program No. HR00112390060, NSF-AI Engage Institute DRL2112635, DARPA Machine Commonsense (MCS) Grant N66001-19-2-4031, and NSF-CAREER Award 1846185. The views contained in this article are those of the authors and not of the funding agency.

\bibliography{custom}

\appendix

\section{Jensen-Shannon Divergence}
\label{app:jsd}
Jensen–Shannon divergence (JSD) is a symmetric measure of the similarity between two probability distributions, defined as the average of the Kullback–Leibler divergences from their mean distribution.
JSD between two probability distribution $P$ and $Q$ is defined as:
\begin{equation*}
    \mathrm{JSD}(P \parallel Q) = \frac{1}{2}\left(\mathrm{KL}(P \parallel M) + \mathrm{KL}(Q \parallel M)\right)
\end{equation*}
where $M = \frac{1}{2}(P + Q)$ is a mixture distribution of $P$ and $Q$ and:
\begin{eqnarray*}
    \mathrm{KL}(P \parallel M) &= \sum_{x}P(x)\log\frac{P(x)}{M(x)} \\
    \mathrm{KL}(Q \parallel M) &= \sum_{x}Q(x)\log\frac{Q(x)}{M(x)}
\end{eqnarray*}

\subsection{JSD Value Trend for Summarization}
\label{app:jsd_trend}
\begin{figure}
    \centering
    \includegraphics[width=1.0\linewidth]{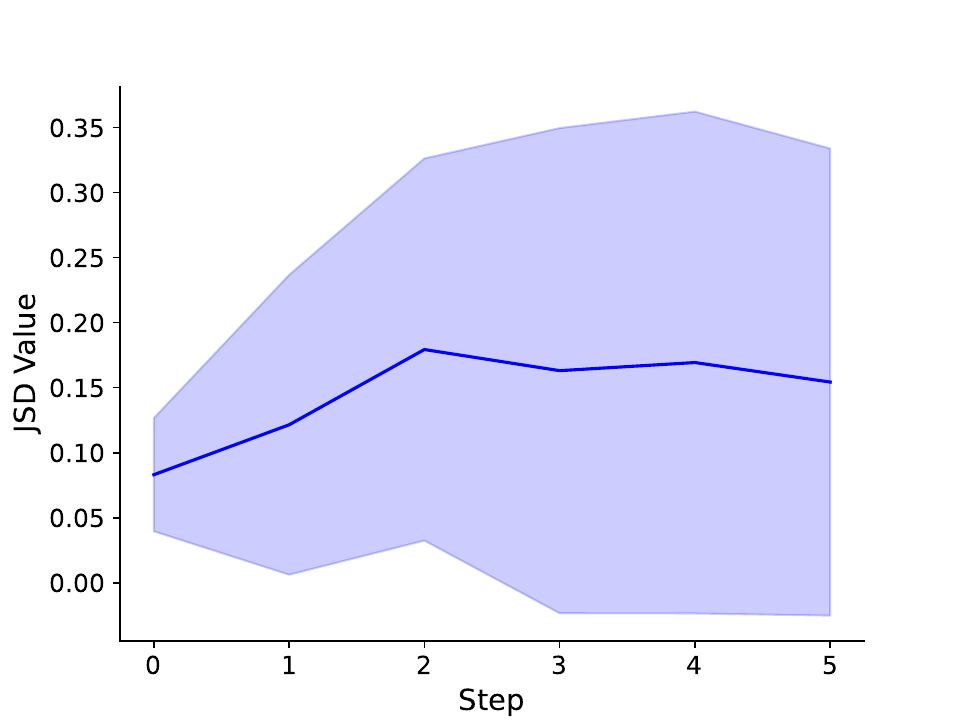}
    \caption{Plot of JSD values of the first 5 decoding steps using Llama3-70B on TofuEval. The JSD values tend to have lower values and variance at the start of decoding.}
    \label{fig:jsd_trend}
\end{figure}

\cref{fig:jsd_trend} illustrates the trend of JSD values over the initial decoding steps when using LLama3-70B on the TofuEval dataset.
We observe that JSD values start relatively low and exhibit less variation or sensitivity in the early steps of decoding.
This may be due to the model's tendency to produce generic, low-information outputs at the start of each sequence.
As the decoding progresses, the JSD values increase and become more sensitive, indicating the dynamic adjustment in \method{} works well.

\section{Dataset Details}
\label{app:data}

We use six question answering datasets and three summarization datasets for evaluation. We also present one example from each dataset, as detailed in \cref{tab:input_example}. For the synthetically generated QA datasets NQ-\textsc{Swap} and NQ-\textsc{Synth}, we provide examples in \cref{tab:nq_swap_example}.
\subsection{QA Datasets}
Some QA datasets, such as NQ, TriviaQA, and HotpotQA, do not have public test sets.
For these we report performance of baselines and \method{} on the dev set. Further, following \citet{shi-etal-2024-trusting}, to expedite inference, we sub-sample datasets where the test sets are very large (>8K instances). 
\begin{itemize}
     \item \textbf{Natural Question} (\textbf{NQ}; \citealp{kwiatkowski-etal-2019-natural}) is a large-scale QA dataset consisting of real user questions issued to Google search, with answers found from Wikipedia. We test on 3231 instances from the NQ validation set, which originally contained 7.83K examples. These instances were selected because they have short answers, making them suitable for evaluating all baselines and our method. 
     \item \textbf{NQ-SWAP} \citep{longpre-etal-2021-entity} introduces synthetic conflicts by swapping entities in the context to challenge the model's ability to manage conflicting information.
     Specifically, \citet{longpre-etal-2021-entity} first identify instances with named entity answers, then substitute mentions of the entity in the gold document with an alternate entity.
     NQ-SWAP consists of 4K instances derived from the NQ dataset.
    \item \textbf{TriviaQA} \citep{joshi-etal-2017-triviaqa} is a realistic QA dataset that includes a wide variety of trivia questions, requiring models to deal with large amounts of text from various sources and handle inference over multiple sentences. We randomly sample 1K instances from the TriviaQA Wiki validation set, which contains a total of 8K examples.
    \item \textbf{PopQA} \citep{mallen-etal-2023-trust} is a dataset designed to test models' performance on questions about long-tail entities. We choose 1.6K instances from the PopQA test set for which we are successfully able to retireive contexts containing the gold answer (c.f. \cref{sec:setup}).
    \item \textbf{HotpotQA} \citep{yang-etal-2018-hotpotqa} is a QA dataset that requires multi-hop reasoning, where the model needs to find and combine information from multiple sources to answer complex questions. We use the entire development set of HotpotQA, consisting of 7.4K instances.
    \item \textbf{TabMWP} \citep{lu2023dynamic} is a dataset focused on open-domain grade-level problems that require mathematical reasoning on both textual and tabular data. We use an official ``lite'' subset of TabMWP called ``test1k'' which contains 1K instances.
\end{itemize}

\subsection{Summarization Datasets}
\begin{itemize}
    \item \textbf{CNN-DM} \citep{see-etal-2017-get} is a widely used dataset for training and evaluating models on abstractive summarization tasks, involving news articles and their summaries. We randomly sample 500 examples from the original 11.5k test set.
    \item \textbf{XSum} \citep{narayan-etal-2018-dont} is an abstractive summarization dataset known for its highly challenging nature, where the goal is to generate concise, one-sentence summaries from longer documents. We used 500 instances from the XSum dataset's 11.3K test set.
    \item \textbf{TofuEval} \citep{tang-etal-2024-tofueval} is a benchmark for evaluating the factual consistency and topic relevance of summaries, especially in scenarios involving dialogue or meeting transcriptions. This benchmark draws 50 test set documents from each of two datasets: MediaSum \citep{zhu-etal-2021-mediasum} and MeetingBank \citep{hu-etal-2023-meetingbank}. For each document, three topics were generated, resulting in a total of 300 topic-focused summaries. Approximately 75\% of the total are main topics that refer to the central information in a document that is under discussion or is presented in the document, and the rest are marginal topics that refer to information in a document that is not the main focus of the document but is still part of the context.
\end{itemize}

\begin{table}[t]
\renewcommand{\arraystretch}{1.1}
\small
\begin{tabular}{p{7.2cm}}
\toprule
\textbf{Question:} \\
How many episodes are in Chicago Fire season 4? \\ \midrule
\multicolumn{1}{c}{\textbf{\textsc{Natural Question}}} \\
\textbf{Original Context:} \\
The fourth season of Chicago Fire contained \textcolor{cyan}{23} episodes. It is an American drama television series with ... \\
\textbf{Original Answer}: \textcolor{cyan}{23} \\ \midrule
\multicolumn{1}{c}{\textbf{\textsc{NQ-Swap}}} \\
\textbf{Substitute Context:} \\
The fourth season of Chicago Fire contained \textcolor{orange}{10} episodes. It is an American drama television series with ... \\ 
\textbf{Substitute Answer}: \textcolor{orange}{10} \\ \midrule
\multicolumn{1}{c}{\textbf{\textsc{NQ-Synth}}} \\
\textbf{Substitute Context:} \\
The fourth season of Chicago Fire contained \textcolor{violet}{22} episodes. It is an American drama television series with ... \\ 
\textbf{Substitute Answer (generated from LLM)}: \textcolor{violet}{22} \\
\bottomrule
\end{tabular}
\caption{Example from \textsc{NQ-Swap} and \textsc{NQ-Synth}. 
A \textcolor{orange}{substitute example} for \textsc{NQ-Swap} is made from the \textcolor{cyan}{original example} by replacing the original answer, \textcolor{cyan}{23}, with a similar but conflicting answer, i.e., \textcolor{orange}{10}. 
A \textcolor{violet}{substitute example} for \textsc{NQ-Synth} is made from the \textcolor{cyan}{original example} by replacing the original answer, \textcolor{cyan}{23}, with one generated by Llama3-70B without context, i.e., \textcolor{violet}{22}. }
\label{tab:nq_swap_example}
\end{table}

\subsection{Licenses}
Datasets are released under the following licenses:
\begin{itemize}
    \item Natural Questions:  Apache-2.0 license
    \item NQ-Swap: MIT license
    \item TriviaQA: Apache-2.0 license
    \item PopQA: MIT license
    \item HotPotQA: Apache-2.0 license
    \item TabMWP: MIT license
    \item CNN-DM: Apache-2.0 license
    \item XSum: MIT license
    \item TofuEval: MIT license
\end{itemize}
The models we use have the following licenses:
\begin{itemize}
    \item Llama 2: custom license \url{https://ai.meta.com/llama/license/}
    \item Llama 3: custom license \url{https://www.llama.com/llama3/license/}
    \item Mistral: Apache-2.0 license
\end{itemize}

\section{Instruction-tuned LLMs Experiments}
\label{app:post-train}

We compare \method{} against the baselines on all datasets using instruction-tuned language models and show the results in \cref{tab:inst_qa_0shot_res}.
We find that \method{} achieves comparable or better performance than all baselines when applied to instruction-tuned models. 

\begin{table*}[t]
\small
\centering
\resizebox{1.0\textwidth}{!}{
\begin{tabular}{llccccccc}
\toprule
\multicolumn{1}{l}{Model} & Decoding & NQ & \textsc{NQ-Swap} & TriviaQA & PopQA & HotpotQA & TabMWP & Avg \\
\midrule
\multirow{3}{*}{Llama2-13B-Chat} & Greedy & 35.75 & 50.24 & 54.40 & \textbf{72.61} & 32.15 & 50.40 & 49.26 \\
 & CAD & \textbf{39.49} & \textbf{71.24} & 59.40 & 68.81 & 30.14 & 48.70 & \textbf{52.96} \\
 & \method{} & 37.08 & 57.69 & \textbf{61.20} & 72.31 & \textbf{32.34} & \textbf{52.10} & 52.12 \\
\midrule
\multirow{3}{*}{Llama3-8B-Inst} & Greedy & \textbf{40.27} & 60.89 & \textbf{64.00} & \textbf{70.89} & \textbf{39.66} & \textbf{68.50} & 57.37 \\
 & CAD & 39.43 & \textbf{71.19} & 52.30 & 70.35 & 37.27 & 63.10 & 55.61 \\
 & \method{} & 39.65 & 67.37 & 61.50 & 70.41 & 39.43 & 66.10 & \textbf{57.41} \\
\midrule
\multirow{3}{*}{Llama3-70B-Inst} & Greedy & 40.82 & 59.16 & 64.10 & 64.41 & 47.70 & 70.40 & 57.77 \\
 & CAD & \textbf{42.31} & \textbf{66.37} & 58.40 & 64.23 & 47.21 & 69.30 & 57.97 \\
 & \method{} & 41.35 & 60.77 & \textbf{64.60} & \textbf{65.78} & \textbf{48.21} & \textbf{71.90} & \textbf{58.77} \\
\midrule
\multirow{3}{*}{Mistral-7B-Inst} & Greedy & \textbf{42.93} & 64.74 & \textbf{77.20} & 76.59 & \textbf{50.26} & \textbf{50.20} & \textbf{60.32} \\
 & CAD & 42.56 & \textbf{67.89} & 71.70 & 74.45 & 47.12 & 46.40 & 58.35 \\
 & \method{} & 42.87 & 63.99 & 75.40 & \textbf{76.89} & 49.49 & 47.30 & 59.32 \\
 \bottomrule
\end{tabular}
}
\caption{Results on QA datasets with different instruction-tuned language models. 
When averaged across datasets, \method{} is better than or comparable to the baselines.}
\label{tab:inst_qa_0shot_res}
\end{table*}

\begin{table*}[t]
\small
\centering
\setlength{\tabcolsep}{2pt}
\begin{tabular}{lccc|ccc|cc}
\toprule
 & \multicolumn{3}{c}{\textbf{CNN-DM}} & \multicolumn{3}{c}{\textbf{XSum}} & \multicolumn{2}{c}{\textbf{TofuEval (AlignScore)}} \\ \cmidrule{2-9} 
\textbf{Decoding} & \textbf{ROUGE-L} & \textbf{BERT-P} & \textbf{AlignScore} & \textbf{ROUGE-L} & \textbf{BERT-P} & \textbf{AlignScore} & \textbf{Overall} & \textbf{Main / Marginal} \\ \midrule
\multicolumn{9}{c}{Llama2-13B} \\ \midrule
Greedy & 23.70 & 94.25 & 87.28 & 13.51 & 93.30 & \textbf{85.23} & 66.11 & 72.51 / 46.23 \\
CAD & \textbf{24.33} & 94.44 & 88.99 & \textbf{14.86} & 93.36 & 82.41 & \textbf{80.39} & \textbf{84.03} / 69.07  \\
COIECD & 20.21 & 88.63 & 75.72 & 13.95 & 89.80 & 70.41 & 62.88 & 68.45 / 45.55 \\
\method{} & 23.93 & \textbf{94.63} & \textbf{91.15} & 14.18 & \textbf{94.04} & 84.33 & \textbf{80.39} & 83.94 / \textbf{69.36} \\ \midrule
\multicolumn{9}{c}{Llama3-8B} \\ \midrule
Greedy & 25.16 & 94.92 & 90.33 & 13.16 & 93.43 & 83.65 & 68.17 & 73.51 / 51.57 \\
CAD & 24.91 & 94.70 & 91.44 & 13.80 & 93.37 & \textbf{86.88} & \textbf{83.40} & \textbf{86.77} / \textbf{72.94} \\
COIECD & 23.60 & 92.01 & 83.92 & 13.65 & 91.40 & 69.47 & 70.07 & 73.65 / 58.94 \\
\method{} & \textbf{25.42} & \textbf{95.09} & \textbf{94.35} & \textbf{13.83} & \textbf{94.02} & 86.78 & 80.62 & 83.24 / 72.46 \\ \midrule
\multicolumn{9}{c}{Llama3-70B} \\ \midrule
Greedy & 24.93 & \textbf{95.41} & 91.44 & 14.36 & 94.05 & 85.28 & 76.66 & 81.64 / 61.19 \\
CAD & 24.76 & 94.45 & 91.01 & 14.59 & 93.65 & 84.34 & 83.93 & 87.26 / 73.58 \\
COIECD & 23.47 & 92.06 & 85.49 & 14.51 & 91.04 & 73.81 & 75.24 & 80.68 / 58.31 \\
\method{} & \textbf{25.42} & 94.91 & \textbf{94.97} & \textbf{14.91} & \textbf{94.29} & \textbf{85.81} & \textbf{85.07} & \textbf{88.06} / \textbf{75.79} \\ \midrule
\multicolumn{9}{c}{Mistral-7B} \\ \midrule
Greedy & 24.59 & 93.57 & 80.80 & 14.07 & 88.56 & 58.76 & 63.07 & 68.62 / 45.79 \\
CAD & 23.72 & 93.22 & 90.61 & 18.20 & 91.54 & 84.94 & 67.64 & 67.55 / \textbf{67.48} \\
COIECD & 23.50 & 92.06 & 83.97 & 17.85 & 89.79 & 69.26 & 65.95 & 70.63 / 51.39 \\
\method{} & \textbf{24.76} & \textbf{94.21} & \textbf{93.05} & \textbf{18.51} & \textbf{92.19} & \textbf{86.79} & \textbf{74.00} & \textbf{77.59} / 62.84 \\ 
\midrule
\multicolumn{9}{c}{Llama3-70B-Instruct } \\ \midrule
Greedy & 24.72 & 90.64 & 88.22 & \textbf{23.19} & 90.80 & 82.40 & 78.56 & 80.18 / 73.52 \\
CAD & 25.17 & \textbf{91.19} & 88.52 & 20.92 & \textbf{91.52} & \textbf{86.54} & 79.86 & 79.55 / \textbf{80.82}  \\
COIECD & 23.85 & 89.84 & 83.88 & 22.41 & 90.61 & 81.42 & 77.54 & 78.69 / 73.97 \\
AdaCAD & \textbf{25.26} & 90.91 & \textbf{88.68} & 21.52 & 91.30 & 85.30 & \textbf{81.16} & \textbf{82.82} / 76.03 \\ \bottomrule
\end{tabular}
\caption{Results on summarization datasets with different LMs. \method{} generally outperforms the baselines across metrics and datasets.}
\label{tab:sum_res_full}
\end{table*}

\section{Results of Different Decoding Methods on CNN-DM}
\label{app:diff_decode}
\cref{tab:diff_decode} shows the results of different base decoding methods on CNN-DM with Llama-70B. 
Here, we see that greedy decoding performs better than Top-$p$ sampling~\cite{Holtzman2020The}, motivating our use of greedy decoding in \cref{tab:sum_res}.

\begin{table}[H]
\small
\centering
\begin{tabular}{lcc}
\toprule
Decoding & ROUGE-L & BERT-P \\ \midrule
Top-$p$ Sampling & 17.48 & 86.79 \\
Greedy Decoding & \textbf{23.47} & \textbf{92.06} \\ \bottomrule
\end{tabular}
\caption{Comparison of greedy decoding and top-$p$ sampling ($p = 0.9$) with Llama3-70B on CNN-DM.}
\label{tab:diff_decode}
\end{table}

\section{Full Results with Different Base LMs on Summarization Tasks}
\label{app:sum_res_full}
\cref{tab:sum_res_full} shows the full results with all base language models on three summarization tasks: CNN-DM, XSum, and TofuEval. \method{} achieves comparable or better performance than all baselines across different LLMs.

\section{Prompts}
\label{app:prompts}

We provide the prompts for pre-trained base language with and without context for both QA and summarization tasks.

\begin{figure}[!h]
\vspace{1em}
\begin{subfigure}{0.5\textwidth}
\centering
\begin{tcolorbox}[
    fontupper=\scriptsize,
    fontlower=\scriptsize,
    colback=white, 
    colframe=gray, 
    title=\textbf{Question Answering}, 
    fonttitle=\bfseries\small, 
    arc=4mm, 
]
\textbf{With Context:}

\{\texttt{context}\}

Using only the references listed above, answer the following question: 

Question: \{\texttt{question}\}

Answer:

\tcblower
\textbf{Without Context:}

Answer the following question: 

Question: \{\texttt{question}\}

Answer:

\end{tcolorbox}
\vspace{1em}
\end{subfigure}
\begin{subfigure}{0.5\textwidth}
\centering
\begin{tcolorbox}[
    fontupper=\scriptsize,
    fontlower=\scriptsize,
    colback=white, 
    colframe=gray, 
    title=\textbf{Summarization - CNN-DM}, 
    fonttitle=\bfseries\small, 
    arc=4mm, 
]
\textbf{With Context:}

Document: \{\texttt{document}\}

Summarize the document in three sentences.

Summary:
\tcblower
\textbf{Without Context:}

Summarize the document in three sentences.

Summary:
\end{tcolorbox}
\end{subfigure}
\end{figure}

\begin{figure}[!h]
\vspace{-7em}
\begin{subfigure}{0.5\textwidth}
\centering
\begin{tcolorbox}[
    fontupper=\scriptsize,
    fontlower=\scriptsize,
    colback=white, 
    colframe=gray, 
    title=\textbf{Summarization - XSum}, 
    fonttitle=\bfseries\small, 
    arc=4mm, 
]
\textbf{With Context:}

Document: \{\texttt{document}\}

Summarize the document in one sentence.

Summary:
\tcblower
\textbf{Without Context:}

Summarize the document in one sentence.

Summary:
\end{tcolorbox}
\vspace{1em}
\end{subfigure}
\begin{subfigure}{0.5\textwidth}
\centering
\begin{tcolorbox}[
    fontupper=\scriptsize,
    fontlower=\scriptsize,
    colback=white, 
    colframe=gray, 
    title=\textbf{Summarization - TofuEval}, 
    fonttitle=\bfseries\small, 
    arc=4mm, 
]
\textbf{With Context:}

Document: \{\texttt{document}\}

Summarize the provided document focusing on ``\{\texttt{topic}\}''. The summary should be less than 50 words in length.

Summary:

\tcblower
\textbf{Without Context:}

Summarize the provided document focusing on ``\{\texttt{topic}\}''. The summary should be less than 50 words in length.

Summary:

\end{tcolorbox}
\end{subfigure}
\end{figure}

\begin{table*}[t]
\small
\centering
\begin{tabular}{p{14cm}}
\toprule
\multicolumn{1}{c}{\textbf{Natural Question}} \\ \midrule
$\boldsymbol{c}$: The second season of the American television drama series Breaking Bad premiered on March 8, 2009 and concluded on May 31, 2009. It consisted of 13 episodes, each running approximately 47 minutes in length ... \\
$\boldsymbol{x}$: How many episodes in season 2 Breaking Bad? \\ \bottomrule\toprule
\multicolumn{1}{c}{\textbf{NQ-SWAP}} \\ \midrule
$\boldsymbol{c}$: The second season of the American television drama series Breaking Bad premiered on March 8, 2009 and concluded on May 31, 2009. It consisted of 27 episodes, each running approximately 47 minutes in length ... \\
$\boldsymbol{x}$: How many episodes in season 2 Breaking Bad? \\ \bottomrule\toprule
\multicolumn{1}{c}{\textbf{TriviaQA}} \\ \midrule
$\boldsymbol{c}$: ... Removal of dental biofilm is important as it may become acidic causing demineralization of the teeth (also known as caries) or harden into calculus (dental) (also known as tartar). Calculus can not be removed through ... \\
$\boldsymbol{x}$: In dentistry, what is the name given to hardened dental plaque? \\ \bottomrule\toprule
\multicolumn{1}{c}{\textbf{PopQA}} \\ \midrule
$\boldsymbol{c}$: The 2012 Uzbekistan First League was the 21st season of 2nd level football in Uzbekistan since 1992. It is split in an Eastern and Western zone, each featuring 12 teams ...\\
$\boldsymbol{x}$: What sport does 2012 Uzbekistan First League play? \\ \bottomrule\toprule
\multicolumn{1}{c}{\textbf{HotpotQA}} \\ \midrule
$\boldsymbol{c}$: <t> Superdrag </t> Superdrag was an American alternative rock band from Knoxville, Tennessee ... \\
<t> Collective Soul </t> Collective Soul is an American rock band originally from Stockbridge, Georgia ... \\
$\boldsymbol{x}$: Are both Superdrag and Collective Soul rock bands? \\ \bottomrule\toprule
\multicolumn{1}{c}{\textbf{TabMWP}} \\ \midrule
$\boldsymbol{c}$: alpaca | \$1,605.00 \\
kinkajou | \$1,837.00 \\
python | \$8,343.00 \\
parrot | \$1,123.00 \\
macaw | \$1,629.00 \\
$\boldsymbol{x}$: Erik has \$7,616.00. How much money will Erik have left if he buys a parrot and a kinkajou? (Unit: \$)\\ \bottomrule\toprule
\multicolumn{1}{c}{\textbf{CNN-DM}} \\ \midrule
$\boldsymbol{c}$: Article: (CNN)Two years ago, the storied Boston Marathon ended in terror and altered the lives of runners, spectators and those who tried to come to their rescue. Just last week, Dzhokhar Tsarnaev was convicted ... \\
$\boldsymbol{x}$: Summarize the article in three sentences. Summary: \\ \bottomrule\toprule
\multicolumn{1}{c}{\textbf{XSum}} \\ \midrule
$\boldsymbol{c}$: You may want to choose another fantasy destination after the British Foreign Office told tourists to be aware that some political demonstrations in the capital, Male, have led to violence. It did add, though, that most trips ... \\
$\boldsymbol{x}$: Summarize the article in one sentence. Summary: \\ \bottomrule\toprule
\multicolumn{1}{c}{\textbf{TofuEval}} \\ \midrule
$\boldsymbol{c}$: Document: DOBBS: General Motors today announced it will offer early retirement buyouts for 113,000 of its employees. Management calls it, ``accelerated attrition''. And it is only the latest sign of the dramatic decline ... \\
$\boldsymbol{x}$: Summarize the provided document focusing on ``Buyouts for General Motors employees''. The summary should be less than 50 words in length. Summary: \\ \bottomrule\toprule
\end{tabular}
\caption{An illustration of input query $\boldsymbol{x}$ and relevant context $\boldsymbol{c}$ for different datasets.}
\label{tab:input_example}
\end{table*}

\end{document}